\documentclass{article}
\usepackage{spconf,amsmath,graphicx}
\usepackage{cite}
\usepackage{amsmath,amssymb,amsfonts}
\usepackage{algorithmic}
\usepackage{graphicx}
\usepackage{booktabs}
\usepackage{multicol}
\usepackage{multirow}
\usepackage{textcomp}
\usepackage{array,xcolor,colortbl}
\usepackage{svg}
\usepackage{hyperref}
\usepackage{xcolor}
\usepackage{stfloats}

\usepackage{enumitem}
\setlist{nosep, leftmargin=14pt}

\usepackage{mwe} 

\title{AMM-Diff: Adaptive Multi-Modality Diffusion Network for Missing Modality Imputation}
\name{Aghiles Kebaili $^1$, Jérôme Lapuyade-Lahorgue $^1$, Pierre Vera $^{1, 2}$ and Su Ruan $^1$}
\address{$^1$ Quantif, University of Rouen-Normandy, Rouen, 76183, France\\
$^2$ CLCC Henri Becquerel, Rouen, 76038, France}

\begin{document}
%
\maketitle
\begin{abstract}
In clinical practice, full imaging is not always feasible, often due to complex acquisition protocols, stringent privacy regulations, or specific clinical needs. However, missing MR modalities pose significant challenges for tasks like brain tumor segmentation, especially in deep learning-based segmentation, as each modality provides complementary information crucial for improving accuracy. A promising solution is missing data imputation, where absent modalities are generated from available ones. While generative models have been widely used for this purpose, most state-of-the-art approaches are limited to single or dual target translations, lacking the adaptability to generate missing modalities based on varying input configurations. To address this, we propose an Adaptive Multi-Modality Diffusion Network (AMM-Diff), a novel diffusion-based generative model capable of handling any number of input modalities and generating the missing ones. We designed an Image-Frequency Fusion Network (IFFN) that learns a unified feature representation through a self-supervised pretext task across the full input modalities and their selected high-frequency Fourier components. The proposed diffusion model leverages this representation, encapsulating prior knowledge of the complete modalities, and combines it with an adaptive reconstruction strategy to achieve missing modality completion. Experimental results on the BraTS 2021 dataset demonstrate the effectiveness of our approach.
\end{abstract}
\section{Introduction} \label{sec:intro}
In recent years, deep learning has demonstrated remarkable efficacy across a range of medical imaging modalities, such as MRI, PET and CT scans. Particularly in the analysis of brain tumors, MRI with its different sequences offers high precision in distinguishing between healthy and pathological tissues \cite{zhou2023literature}. However, one of the persistent challenges in medical imaging is the absence of certain modalities, often due to complex acquisition protocols, strict privacy regulations, or specific clinical needs limiting their availability. This is particularly critical as each modality offers distinct medical insights, with substantial variation in how pathological tissues are represented, particularly in multimodal settings where the interdependent semantic relationships between these modalities are crucial. To address this issue, missing data imputation comes as a promising solution by predicting the missing modalities based on the available ones, capitalizing on the complementary information inherent in the existing modalities and cross-modal relationships to synthesize the missing ones.

\begin{figure*}[t]
\begin{center}
\includegraphics[width=16cm]{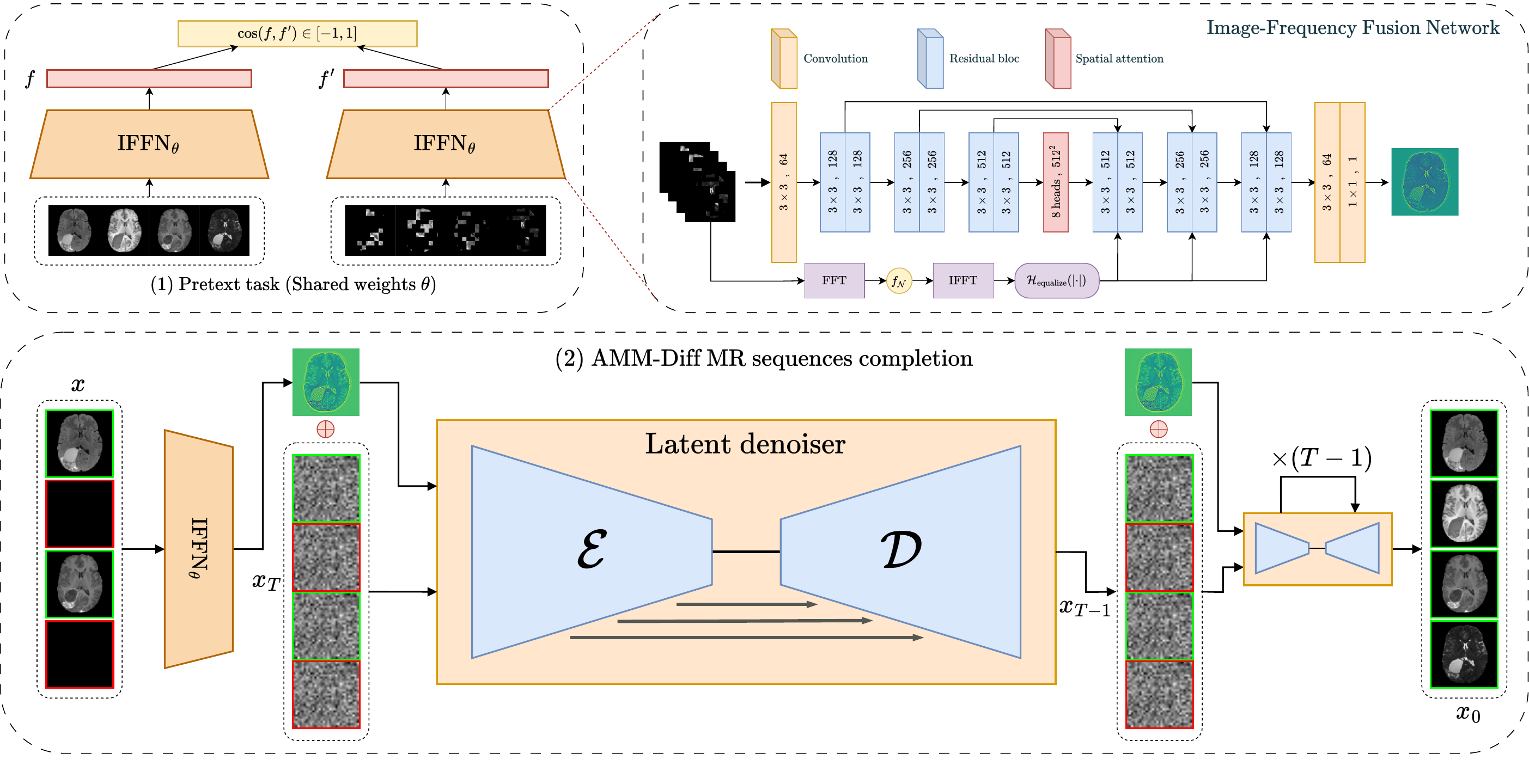}
\caption{Architecture of the proposed AMM-Diff. (1) Self-supervised pretraining of the Siamese IFFN using a cosine similarity loss. (2) Missing sequence completion. Green-bordered images represent inputs and red-bordered images represent the missing sequences. The input configuration here is entirely random. See section \ref{sec:IFFN} and \ref{sec:diffusion} for more details.}\label{fig:architecture}
\end{center}
\end{figure*}
Data imputation predominantly relies on advanced generative models for their capacity to capture the intricate dependencies between different modalities. Extensive research has explored conditional generative frameworks for cross-modal synthesis \cite{kebaili2023deep}, however, most existing state-of-the-art approaches are limited to fixed, single or dual target predictions \cite{meng2022novel,xing2024cross}, which, while effective at exploiting multi-source inputs, inherently limits the model’s adaptability to handle variable configurations of missing modalities. Conditional Generative Adversarial Networks (cGANs) \cite{goodfellow2014generative} have been extensively applied in medical imaging due to their capacity to generate high-quality images. However, despite their success, GANs are often hindered by inherent limitations, 
such as unstable training and convergence issues \cite{kebaili2023deep}. To address these challenges, Variational Autoencoders (VAEs) \cite{kingma2013auto} have been proposed as a more stable alternative, offering improved mode coverage. Yet, VAEs introduce their own drawbacks, particularly the generation of blurry and less detailed images. More recently, the rise of diffusion models \cite{ho2020denoising} has presented a promising solution for high-quality image generation, and gained traction in various medical imaging tasks , including semantic segmentation and image-to-image translation \cite{kazerouni2023diffusion}. Their efficacy lies in their ability to synthesize images with exceptional realism and detail, outperforming traditional methods like VAEs and GANs \cite{dhariwal2021diffusion}, while ensuring broader mode coverage and improved reconstruction accuracy. Recently, Diffusion Models have emerged as a promising solution for synthesizing high-quality images and can be leveraged to enhance data imputation across various modalities.

In this study, we propose a novel diffusion-based generative model named AMM-Diff designed for adaptive missing data completion in multimodal settings. Our architecture is capable of handling any input configuration from a set of input modalities and generating the missing ones through an adaptive reconstruction strategy. We introduce a Siamese Image-Frequency Fusion Network (IFFN), pretrained using a similarity-based pretext task to consolidate spatial feature maps and capture essential inter-modality relationships across input sequences. Inspired by techniques from natural language processing \cite{devlin2018bert}, we employ a masked image modeling strategy, allowing the network to learn from incomplete data. This results in a unified feature representation that encapsulates key features and correlations from the complete input images. The learned representation is then used as input to the translation diffusion model in the downstream task, where it acts as a decoder to synthesize the missing sequences. In this paper, we focus on multi-sequence MRIs, however, our architecture can be extended to other cross-modal translations, such as CT or PET scans. We evaluate the performance of AMM-Diff both quantitatively and qualitatively on the BraTS 2021 dataset. Our main contributions are:
\begin{itemize}
    \item A two-branch Siamese IFFN is designed for learning a spatial and spectral feature respresention through a self-supervised similarity-based pretext task.
    \item A new architecture leveraging the IFFN is proposed for adaptive missing MR sequence completion, capable of handling varying input configurations and adapting to available MR sequences for generating the missing ones.
    \item Quantitative and qualitative evaluation demonstrating the capability of our approach to maintain critical inter-modality information while synthesizing high-quality MR sequences.
\end{itemize}


\begin{table*}[bp]
\centering
\resizebox{\textwidth}{!}{%
\begin{tabular}{@{}lcccccccccccccccc@{}}
\toprule
\multirow{2}{*}{Methods} & \multicolumn{4}{c}{FLAIR} & \multicolumn{4}{c}{T1} & \multicolumn{4}{c}{T1CE} & \multicolumn{4}{c}{T2}\\ 
\cmidrule(lr){2-5} \cmidrule(lr){6-9} \cmidrule(l){10-13} \cmidrule(l){14-17}
& MSE $\downarrow$ & PSNR $\uparrow$ & SSIM $\uparrow$ & LPIPS $\downarrow$ & MSE $\downarrow$ & PSNR $\uparrow$ & SSIM $\uparrow$ & LPIPS $\downarrow$ & MSE $\downarrow$ & PSNR $\uparrow$ & SSIM $\uparrow$ & LPIPS $\downarrow$ & MSE $\downarrow$ & PSNR $\uparrow$ & SSIM $\uparrow$ & LPIPS $\downarrow$ \\ 
\cmidrule(r){1-1} \cmidrule(lr){2-5} \cmidrule(lr){6-9} \cmidrule(l){10-13} \cmidrule(l){14-17}
Pix2Pix                         & 0.0019 & 28.080 & 0.6943 & 0.1299
                                & 0.0024 & 30.301 & 0.8806 & 0.0566
                                & 0.0011 & 30.421 & 0.7198 & 0.1306 
                                & 0.0013 & 30.014 & 0.7385 & 0.1110 \\

UMM-CSGM                     & 0.0019 & 29.416 & 0.9107 & 0.0324
                                & 0.0027 & 28.944 & 0.9256 & 0.0191 
                                & 0.0014 & 31.391 & 0.9112 & 0.0309 
                                & 0.0013 & 30.250 & 0.9272 & 0.0268 \\

AMM-Diff w/ U-Net               & 0.0104 &  22.090 & 0.7551 & 0.0609 
                                & 0.0412 &  14.930 & 0.7144 & 0.0821 
                                & 0.0044 &  25.697 & 0.8596 & 0.0449 
                                & 0.0053 &  23.554 & 0.8491 & 0.0410 \\
                                
AMM-Diff w/ IFFN                & 0.0015 &  29.348 & 0.8804 & 0.0321 
                                & 0.0022 & 28.920 & 0.9246 & 0.0215 
                                & 0.0010 & 31.577 & 0.8889 & 0.0319 
                                & 0.0012 & 30.050 & 0.9297 & 0.0252 \\

AMM-Diff w/ Pretrained IFFN      & \textbf{0.0012} & \textbf{30.707} & \textbf{0.9130} & \textbf{0.0283}
                                & \textbf{0.0017} & \textbf{31.514} & \textbf{0.9548} & \textbf{0.0155} 
                                & \textbf{0.0008} & \textbf{32.631} & \textbf{0.9311} & \textbf{0.0265} 
                                & \textbf{0.0010} & \textbf{32.348} & \textbf{0.9409} & \textbf{0.0235} \\
                                
\bottomrule
\end{tabular}%
}
\caption{Quantitative performance of the proposed generative models on the BRATS datasets.\label{tab:visual_quality}}
\end{table*}

\section{Method}
\subsection{Diffusion models}
Diffusion models \cite{ho2020denoising} approximate data distributions by combining a series of simpler ones through a forward-and-backward stochastic process inspired by nonequilibrium thermodynamics. This methodology enables them to effectively capture complex, high-dimensional structures and generate samples that accurately represent the underlying data. The forward process gradually adds Gaussian noise to an initial data sample $x_0 \sim q(x_0)$, using a variance scheduler $\beta_1, \dots, \beta_T$. At each time step $t$, this process is described by:

\begin{equation}
    q(x_t|x_{t-1}) = \mathcal{N}(x_t; \sqrt{1 - \beta_t} x_{t-1}, \beta_t \mathbf{I})
\end{equation}

The backward process aims to reverse this, beginning with a noisy sample $x_T \sim \mathcal{N}(\mathbf{0}, \mathbf{I})$ and learning to denoise it step by step back into the initial data distribution $q(x_0)$. This reverse step is modeled as:

\begin{equation}
    p_\theta(x_{t-1}|x_t) = \mathcal{N}(x_{t-1}; \mu_\theta(x_t, t); \sigma^2_t)
\end{equation}

During training, the model attempts to predict the added noise $\epsilon$ during each forward step. The parameters $\theta$ are learned through maximum likelihood estimation, analogous to variational inference in VAEs \cite{kingma2013auto}, by maximizing the evidence lower bound (ELBO). The loss is simplified to a mean squared error between predicted and true noise as $|\hat{\epsilon} - \epsilon|^2_2$.

\subsection{Image-Frequency Fusion Network} \label{sec:IFFN}
The proposed IFFN is designed to handle multimodal inputs adaptively by mapping different modality combinations into a unified feature representation that encapsulates essential inter-modality relationships. The IFFN becomes more resilient and invariant to missing modalities, as it can effectively handle the absence of certain MR sequences and still extract meaningful and complementary inter-modal features. As shown in Figure \ref{fig:architecture}, IFFN architecture has a low downscale factor $f = 2$ to preserve maximum spatial information, and fuses the input modalities into a single unified feature map through a pixel-wise translation. The IFFN also employs a spectral fusion mechanism, where the input images are transformed into the Fourier frequency domain, and extracts high-frequency features according to the following equation:

\begin{equation}
    x_{\text{HL}} = \mathcal{H}_{\text{eq}} \left(\left| \mathcal{F}^{-1}\left(\mathcal{F}(x) \cdot f_\mathcal{N}\right)\right|\right)
\end{equation}

Here, $\mathcal{H}_{\text{eq}}$ is the histogram equalization function, $\mathcal{F}$ represents the Fourier transform, and $f_{\mathcal{N}}$ is a smooth high-pass filter that retains essential high-frequency components. These extracted features are integrated into the late layers of the network, helping to capture the overall anatomical structure of the brain and compensating for missing gradient information in absent modalities. This also improves the model’s ability to accurately delineate tumor boundaries.

The model undergoes a two-phase training process: \textbf{(1) Masked Image Modeling:} We employ a similarity-based pretext task using masked image modeling, where a two-branch Siamese IFFN is trained on all modalities. The model minimizes a cosine similarity loss between patch-masked inputs and their corresponding complete multimodal counterparts. This enables the IFFN to learn the underlying correlations between different modalities, which will be distilled as prior knowledge in the case of missing modalities. \textbf{(2) Translation downstream task:} The IFFN is fine-tuned alongside the diffusion model, updating its weights via backpropagation of the diffusion model's loss in an end-to-end manner, similar to the approach in \cite{prabhudesai2024test}. This allows the unified feature representation to continue adapting to the task-specific requirements of the diffusion model.

\subsection{Adaptive MR sequence completion}\label{sec:diffusion}
The second sub-network of the AMM-Diff is the translation diffusion model, which serves as a decoder to synthesize missing MRI modalities based on the unified representation generated by the IFFN. This unified representation addresses the challenge of dynamically varying input modalities, allowing missing sequences to be substituted with blank images. However, decoding these inputs is complex due to the technical constraint that convolutional layers cannot output variable numbers of sequences, as each filter is specialized for specific modalities. To overcome this, we propose an adaptive reconstruction strategy, where the output is fixed to the total number of sequences, and the model reconstructs even the available modalities. This approach ensures that all possible sequences are generated regardless of the input configuration. Additionally, reconstructing the available modalities during the synthesis process helps maintain structural and anatomical coherence across all outputs, reinforcing the model’s understanding of the data distribution. Together, the IFFN and the translation diffusion model form an encoder-decoder framework, where the IFFN encodes a common representation and the diffusion model decodes it to reconstruct both missing and present sequences.

\section{Experimentations}
\subsection{Dataset}
We evaluate the efficacy of our proposed method using the publicly available dataset: BRAin Tumor Segmentation (BRATS2021) \cite{baid2021rsna} provides multi-modal MRIs with a volume shape of 240$\times$240$\times$155 and a voxel resolution of 1$\times$1$\times$1 $mm^3$. In our experiments, we consider all of the four MR sequences: FLAIR, T1, T2 and T1CE. Images are resized to $128 \times 128$ for training time purposes. We respectively use 60\%, 20\%, and 20\% as train, validation and test splits from the initial 1251 available subjects.

\begin{figure}[t]
\begin{center}
\includegraphics[width=\columnwidth]{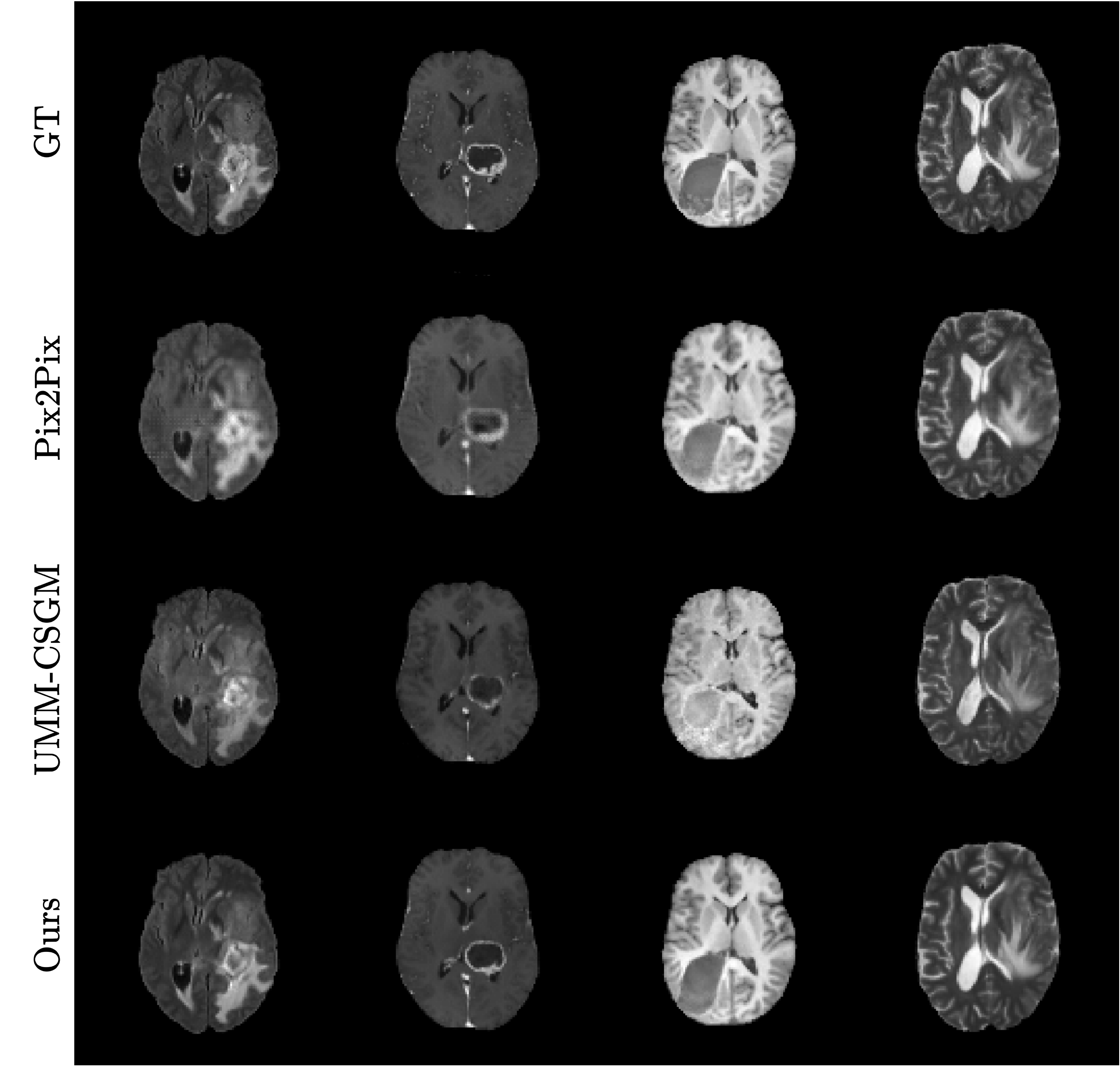}
\caption{Comparison of translated MR sequences across different methods. Columns represent FLAIR, T1CE, T1, and T2 sequences (left to right).}\label{fig:translation_comparison}
\end{center}
\end{figure}

\subsection{Quantitative results}
Table \ref{tab:visual_quality} presents a quantitative comparison of our method with two state-of-the-art techniques: (1) Pix2Pix \cite{isola2017image}, a U-Net-based cGAN widely used for image-to-image translation, and (2) UMM-CSGM \cite{meng2022novel}, a conditional diffusion-based model. We evaluate the methods using pixel-wise, structural, and perceptual metrics to provide a comprehensive analysis. Unlike the comparative methods, which require training a separate model for each missing MR sequence, our approach adaptively generates multiple missing sequences using a single model instance. Although further experiments could have explored 10 additional configurations, we limited our tests to single-target comparisons due to the 4-page limit. In this setup, with three available input modalities, we generate only one missing modality. The results show the superiority of our method across all metrics. While UMM-CSGM offers competitive results, it is specialized for a single sequence, which simplifies the task and reduces model complexity. Our architecture proposes high-quality images despite the increased complexity of generating multiple missing sequences using a shared model. This highlights the strength of our unified feature representation, where the IFFN effectively distills prior knowledge to recover missing modalities, thanks to contrastive pretraining. Acting as an encoder-decoder pair, the IFFN also simplifies the translation task for the translation diffusion model. This is further supported by ablation studies, confirming the IFFN’s ability to capture critical features for missing sequences.

\subsection{Qualitative results}\label{sec:qualitative}
\begin{figure}[t]
\begin{center}
\includegraphics[width=8.2cm]{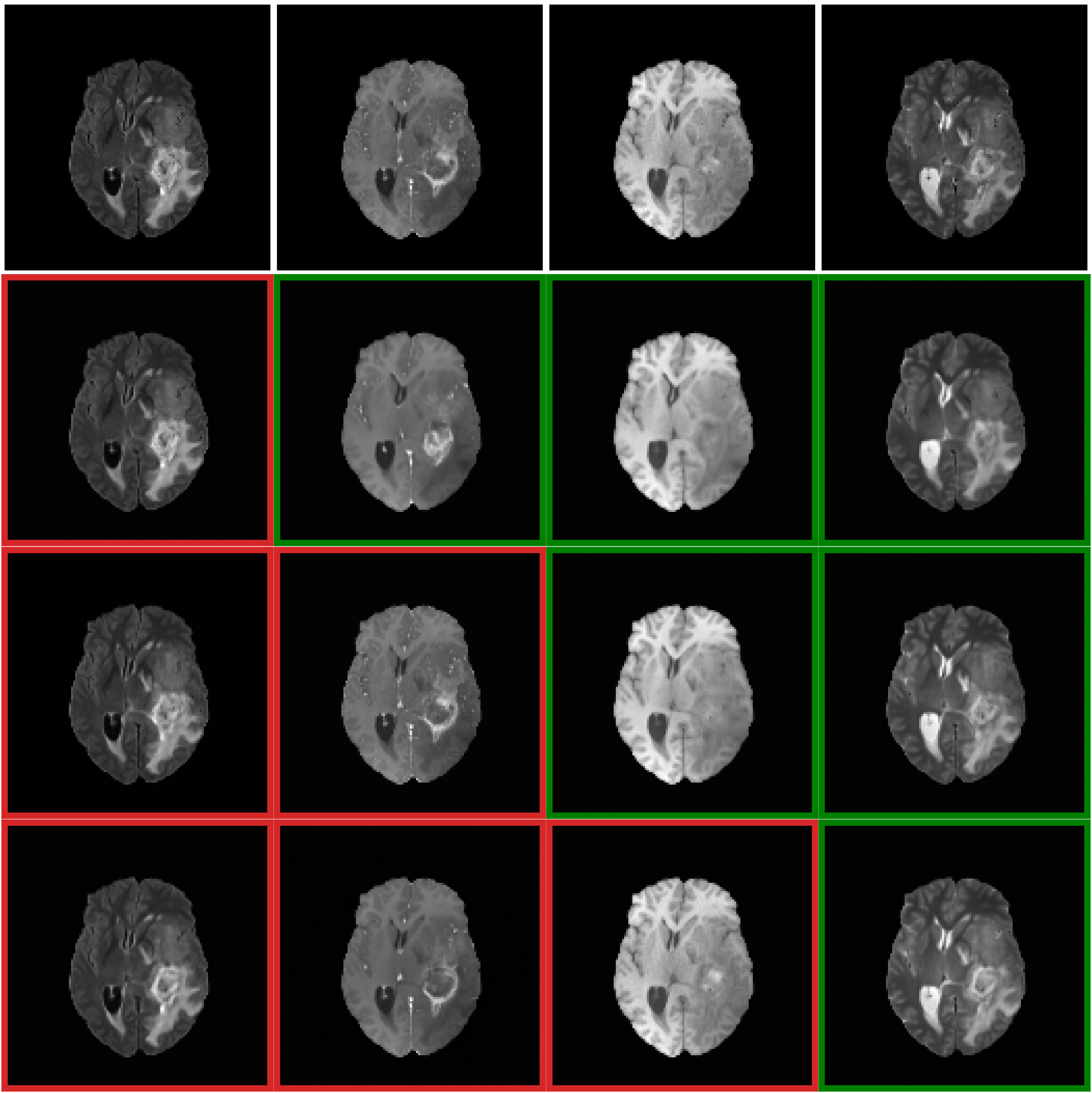}
\caption{Illustration of adaptive translations from various input configurations. Red-bordered images represent inputs sequences and green-bordered images represent the predicted missing ones. See section \ref{sec:qualitative}.}\label{fig:adaptive_comparison}
\end{center}
\end{figure}

Figure \ref{fig:translation_comparison} shows qualitative comparisons of missing sequence translations generated by our method and the two other comparative methods. Our approach demonstrates more realistic and fine-grained details in the overall brain structure, with particularly precise tumor boundary representation compared to UMM-CSGM. The textural contrast between gray and white matter is also better preserved. Pix2Pix, on the other hand, exhibits missing fine-grained details and limited brain structure clarity. Each translation by our model was achieved using the same instance, underscoring its adaptability for dynamically generating missing sequences with high fidelity. In Figure \ref{fig:adaptive_comparison}, we present adaptive translations from various input configurations. The first row shows ground truth, while red and green borders highlight the input and synthesized sequences, respectively. A clear trend is observed: the more input sequences available, the richer the IFFN’s unified representation, particularly improving tumor delineation in T2 sequences from the second-to-last rows. However, when fewer input modalities are provided, there is a tendency for random interpolation of tumor regions, especially without key modality pairs like FLAIR/T2 or T1/T1CE. This is visible in the second row’s T1CE, which improves as more input sequences are added. Despite fewer inputs, our method reliably captures enough information to translate missing sequences, illustrating the robustness of the IFFN. Additionally, our approach, treating 3D images as a sequence of 2D slices, can easily extend to 3D image translation by processing each slice and reconstructing the full 3D volume.

\section{Conclusion}
In this paper, we introduce the AMM-Diff for adaptive imputation of missing MR sequences. Using the proposed IFFN, which employs a self-supervised masked image modeling approach to build a unified feature representation, our method generates high-quality missing sequences through a diffusion model. These synthesized sequences can then be applied to tasks such as brain tumor segmentation and prediction, ultimately improving clinical decision-making and patient outcomes.

\bibliographystyle{IEEEbib}
\bibliography{main}

\end{document}